\pdfoutput=1

\documentclass[11pt]{article}

\usepackage{ACL2023}

\usepackage{times}
\usepackage{latexsym}

\usepackage[T1]{fontenc}

\usepackage[utf8]{inputenc}

\usepackage{microtype}

\usepackage{inconsolata}

\usepackage{booktabs}
\usepackage{amsmath}
\usepackage{multirow}
\usepackage{graphicx}
\usepackage{adjustbox}
\usepackage{amssymb}

\usepackage[normalem]{ulem}

\title{Text-Guided Alternative Image Clustering}

\author{First Author \\
  Affiliation / Address line 1 \\
  Affiliation / Address line 2 \\
  Affiliation / Address line 3 \\
  \texttt{email@domain} \\\And
  Second Author \\
  Affiliation / Address line 1 \\
  Affiliation / Address line 2 \\
  Affiliation / Address line 3 \\
  \texttt{email@domain} \\}

\author{Andreas Stephan$^{1,2,6}$, Lukas Miklautz$^{1}$, Collin Leiber$^{4,5}$,  \\ \bf Pedro Henrique Luz de Araujo$^{1,2}$,
\bf Dominik Répás$^1$,  Claudia Plant$^1$ \and Benjamin Roth$^{1,3}$ \\
$^1$ Faculty of Computer Science, University of Vienna, Austria \\ 
$^2$ UniVie Doctoral School Computer Science, University of Vienna, Austria \\
$^3$ Faculty of Philological and Cultural Studies, University of Vienna, Austria \\
$^4$ LMU Munich, Germany \\
$^5$ Munich Center for Machine Learning, Munich, Germany \\
\texttt{$^6${andreas.stephan}@univie.ac.at}}

\begin{document}
\maketitle

\begin{abstract}
Traditional image clustering techniques only find a single grouping within visual data. 
In particular, they do not provide a possibility to explicitly define multiple types of clustering. 
This work explores the potential of large vision-language models to facilitate alternative image clustering. 
We propose Text-Guided Alternative Image Consensus Clustering (TGAICC), a novel approach that leverages user-specified interests via prompts to guide the discovery of diverse clusterings.  
To achieve this, it generates a clustering for each prompt, groups them using hierarchical clustering, and then aggregates them using consensus clustering. 
TGAICC outperforms image- and text-based baselines on four alternative image clustering benchmark datasets. Furthermore, using count-based word statistics, we are able to obtain text-based explanations of the alternative clusterings. In conclusion, our research illustrates how contemporary large vision-language models can transform explanatory data analysis, enabling the generation of insightful, customizable, and diverse image clusterings. \footnote{Code available at \url{https://github.com/AndSt/alternative_image_clustering}.}
\end{abstract} 


\section{Introduction}

\begin{figure}
    \centering
    \includegraphics[width=0.9\columnwidth]{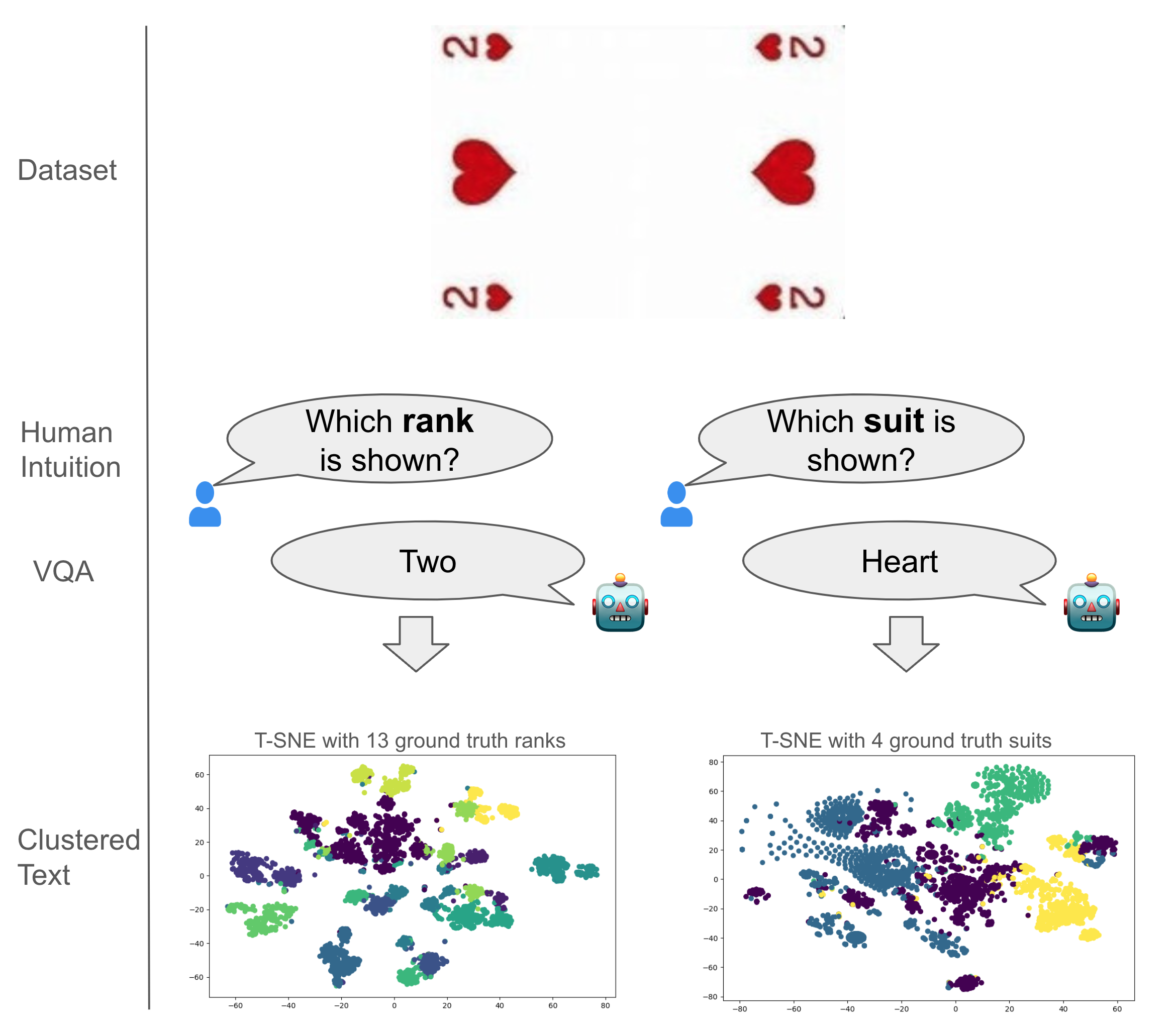}
    \caption{Assume we have an image of a card depicting a ``heart two''. Given two different user queries, the VQA model gives different responses. Clustering the generated text based on different prompts, results in alternative clusterings, satisfying different needs. The colors in the figure represent the ground truths of ``rank'' and ``suit'' for different generated texts.}
    \label{fig:figure_1}
\end{figure}

Exploratory data analysis (EDA) serves a crucial function in the comprehension and analysis of data \cite{tukey1970exploratory}. Clustering arises as a cornerstone EDA methodology, facilitating the grouping of similar data points into coherent groups. A dataset of images, for example, can be clustered based on semantic similarities between the shown objects.
Nevertheless, within applied contexts, variations in user requirements or foci demand distinct clustering formations. One might, for instance, cluster a dataset of cards by rank or by suit (see Figure \ref{fig:figure_1}).
In such circumstances, it is advantageous to derive multifaceted insights into a dataset from diverse perspectives.

Current approaches in alternative image clustering either rely on image-based features \cite{nrkmeans, Miklautz_Mautz_Altinigneli_Boehm_Plant_2020} or utilize text through image-text bi-encoders, often with architectures resembling CLIP \cite{pmlr-v139-radford21a, yao2024multi}.
These methods, while powerful, neglect the rich insights that can be extracted by models explicitly trained to retrieve specific aspects of information from images using text (e.g., visual question answering (VQA) models \cite{Antol_2015_ICCV}). 
Stephan et al. (2024) demonstrate the effectiveness of using generated text descriptions to improve standard image clustering tasks, i.e. in scenarios where a single clustering structure is expected.

We aim to use image-to-text models to obtain alternative clusterings. 
By encoding visual content into text, similarity dimensions beyond the visual features can be explored, potentially uncovering interpretable relationships. 
We introduce \textit{TGAICC (Text-Guided Alternative Image Consensus Clustering)}, a clustering method that uses the generations of multiple image-to-text models to obtain alternative clusterings.
TGAICC incorporates VQA models to generate multiple textual descriptions of images and then clusters the images based on the generated natural language descriptions. We identify similar clusterings using their mutual information, group them using hierarchical clustering and aggregate them using consensus clustering to form refined, alternative clusterings.



Our experimental setup employs four widely used alternative image clustering datasets, each possessing two or three ground truth labelings (e.g., playing cards clustered by rank or suit).  
We compare TGAICC against baselines for alternative clustering using image-only features and baselines that make use of the generated text.
Our experiments demonstrate the following key findings: methods clustering the generated text outperform methods based on image features on these alternative clustering datasets, underscoring the power of textual representations in capturing diverse aspects of similarity. 
Further, TGAICC, on average, achieves superior results across the evaluated datasets and metrics when compared to all other methods, highlighting the effectiveness of our framework in leveraging image-to-text models to uncover alternative and insightful clusterings.
Lastly, we can better interpret the clusterings by explaining the content using word statistics. Our case study on the cards dataset shows that text provides an opportunity to obtain an informative overview of the data.

In summary, our research provides the following contributions:

\begin{enumerate}
   \item We introduce a prompt-based setup to obtain alternative image clusterings.
    \item We introduce TGAICC, a method that combines ideas from multi-modality, hierarchical clustering, and consensus clustering to obtain alternative clusterings.
    \item Our experiments on four common alternative image clustering datasets show that our TGAICC outperforms our baselines
    \item Our methodology enables the ability to generate textual cluster explanations, offering a clear overview of the unique content captured within each alternative clustering.
\end{enumerate}

\section{Related Work}

This work builds upon image clustering, consensus clustering, and alternative clustering approaches. We provide a brief overview of these relevant areas and describe the necessary background.

\subsection{Image Clustering}

Research in image clustering has addressed several standard issues, and a variety of techniques have been developed to tackle them. \citep{clusteringsurvey} provide a survey on clustering approaches.  Classic approaches like k-means \cite{kmeans} have demonstrated effectiveness but often struggle with complex or high-dimensional image data. To address these limitations, more recent work has explored deep clustering methods such as DEC \cite{dec} and IDEC \cite{idec}. 
In addition to these core techniques, representation learning and more specifically, self-supervised learning \cite{surveycontrastiveselfsupervised} has emerged as a vital aspect of image clustering \cite{lehner2023maect, Adaloglou_2023_BMVC}. 
In Contrastive Clustering \cite{li2021contrastive}, the authors use one loss contrasting image features and another loss contrasting clustering features, i.e., the predicted cluster of two augmentations of the same image.
A different approach is used in Text-Guided Image Clustering \cite{stephan-etal-2024-text}. This paradigm leverages image-to-text models and subsequently cluster text. The observation that text often outperforms image-based features motivates this work.

\subsection{Consensus Clustering}

Variability in clustering results arises from different clustering algorithms or variations in their initializations. Given that different clusterings potentially reveal different insights (e.g., accurately identifying a cluster representing "hearts"). 
Consensus clustering methods aim to aggregate results from multiple base clustering algorithms to produce a more robust and stable final clustering. The problem was formalized by \citep{clusterensembles} and the authors introduce the Cluster-based Similarity Partitioning Algorithm (CSPA), HyperGraph Partitioning Algorithm (HGPA), and the Meta-CLustering Algorithm (MCLA). All three employ clustering similarity functions, e.g. Normalized Mutual Information (NMI), to construct a similarity graph and use graph theory to obtain a consensus clustering. In \citep{nmfconsensus}, non-negative matrix factorization (NMF) is used to obtain a consensus clustering.
The Hybrid Bipartite Graph Formulation (HBGF) \cite{10.1145/1015330.1015414} employs a bipartite graph representation. In \citep{deccs}, the authors introduce DECCS, a deep learning-based consensus method, which learns a representation on which heterogeneous clustering algorithms share a consensus on the obtained clusterings. 

\subsection{Alternative Clustering}

Clustering methods usually focus on finding a single optimal clustering solution. Motivated by the fact that there may be multiple meaningful ways to group data points, alternative clustering approaches aim to uncover multiple, diverse clustering structures within the same data \cite{multiple_clusterings_survey, alternative_clustering_survey}. 

\citet{orth} first apply a traditional clustering algorithm and then transform the dataset into a feature space orthogonal to the current clustering. Two strategies are proposed: orthogonal clustering (orth1) and clustering in orthogonal subspaces (orth2). 
In contrast, Non-redundant K-means (Nr-Kmeans) \cite{nrkmeans} simultaneously identifies multiple clusterings within a dataset by iteratively rotating the feature space and assigning features to specific clusterings.
ENRC \cite{Miklautz_Mautz_Altinigneli_Boehm_Plant_2020} is a deep non-redundant clustering method that learns multiple clusterings from a dataset by (soft-)assigning each dimension of the embeddings space to a clustering.
In \citep{kwon2024image}, the authors provide initial text criteria, e.g., suits and ranks, and use image-to-text models to extract information, and then GPT-4 to obtain cluster names and classify images into clusters. Thus, this approach is expensive.
In concurrent work, \citep{yao2024multi} use GPT-4 to generate cluster name candidates and contrastively fine-tune CLIP \cite{pmlr-v139-radford21a}.

\subsection{Image-To-Text Models}

Recently, the development of multimodal models has seen rapid advancement. 
Image-to-text models, in particular, learn to associate visual content with corresponding textual descriptions, which is useful for, e.g., visual question-answering (VQA) \cite{yin2023survey, Antol_2015_ICCV}.

Flamingo \cite{NEURIPS2022_960a172b} allows interleaving images and text by using Perceiver Resamplers on top of pre-trained models.
BLIP and BLIP2 \cite{pmlr-v162-li22n, li2023blip2} employ a frozen image encoder along with a frozen LLM to generate text. LLaVA and LLaVA-NeXT\cite{liu2023llava, liu2023improvedllava} convert image patches into token embeddings using a fixed Vision Transformer encoder followed by a trained MLP. These tokens then become the input for the LLM, enhancing the descriptive results.

In this work, we use LLaVA to extract relevant information from images. More specifically, we frame the image-to-text generation as a VQA task: we prompt LLaVA with an image and corresponding questions about it to generate natural language descriptions of the image.



\begin{figure*}
    \centering
    \includegraphics[width=\textwidth]{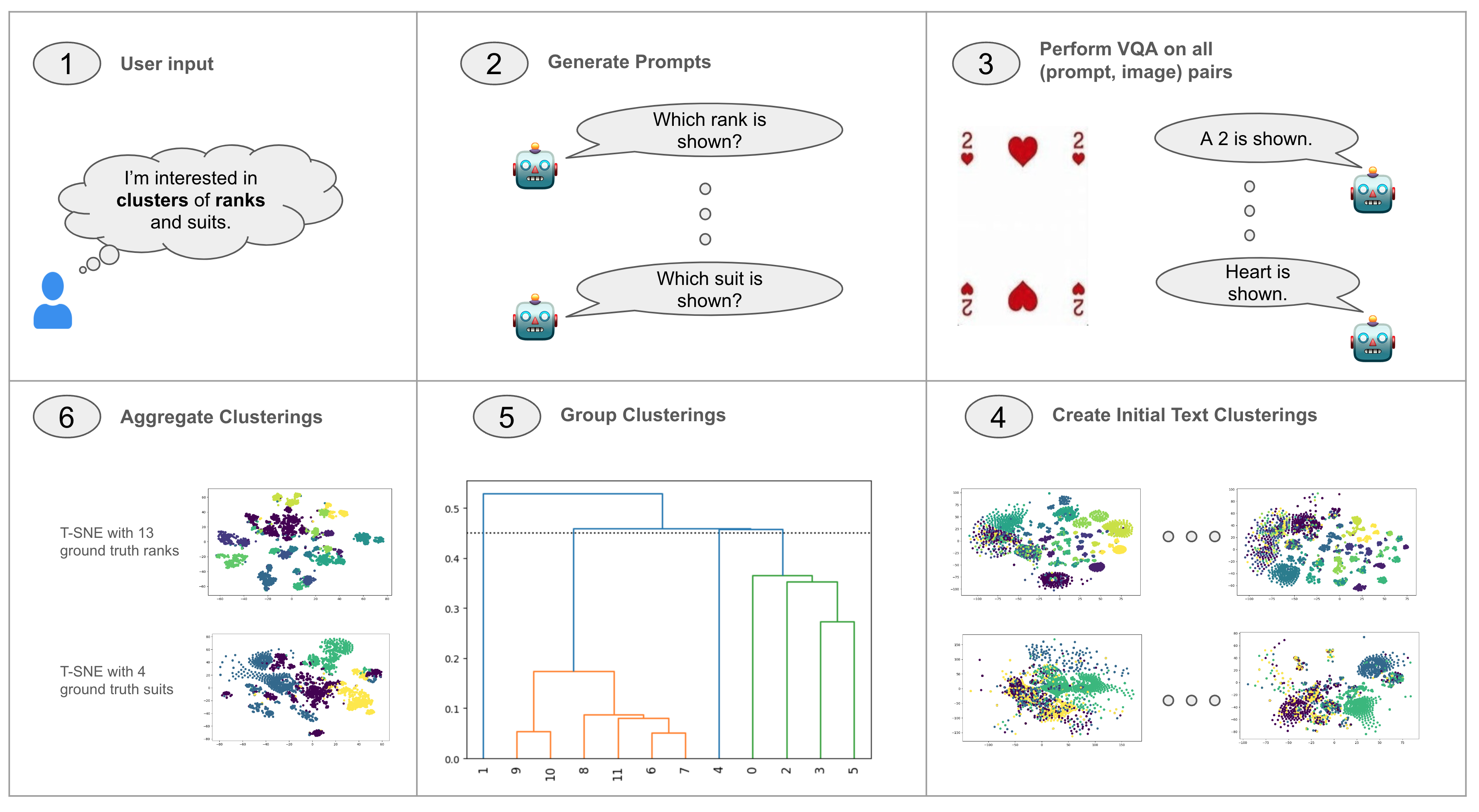}
    \caption{An overview of our methodology. In 1) a user provides text, indicating his interest in the data. In 2) a LLM generates a set of prompts tailored to extract specific information from images, and in 3) VQA is performed for each prompt on each data sample. In 4) the texts generated per prompt are clustered (colors resemble rank and suit ground truth). In 5) a hierarchy of similar clusterings is built. Based on a threshold (dotted line), multiple groups of clusterings (green and orange) are identified and in 6) aggregated to obtain the final alternative clusterings.}
    \label{fig:main_figure}
\end{figure*}
\section{TGAICC}

We use image-to-text models, specifically models that are able to describe specific aspects of information from images in order to obtain different clusterings. Thus, we design prompts to perform VQA. It is well known \cite{bach2022promptsource,sclar2024quantifying} that responses to seemingly semantically equal prompts might vary heavily. Thus, we use multiple formulations of each prompt and aggregate their clusterings afterward. 
Figure \ref{fig:main_figure} gives an overview of the process.

\textbf{Setup.} The input to TGAICC is a dataset of $k$ datapoints, $t$ initial prompts, and, as common in the alternative-clustering literature, the number of clusters in the ground truth clusterings $\{z_1, \dots, z_t\}, z_i \in \mathbf{N}$. 
E.g., $\{2, 4\}$ means the algorithm should return one clustering with $2$ and one clustering with $4$ clusters. 
Note that the difference to the traditional alternative clustering setup is that we give additional initial prompts.
The output is comprised of $t$ clusterings where the $k$ data points are grouped into $z_1, \dots, z_t$ clusters.

\subsection{Initialization}

The following two paragraphs describe the steps 1-4 in Figure \ref{fig:main_figure}.

\textbf{Prompt Design}  We write a query and ask GPT-4 \cite{openai2024gpt4} to automatically generate additional questions. The specific prompt is '\textit{Generate three diverse paraphrases for the following question: \{initial question\}}'.
Further, we generate a variation of each prompt by appending the directive "Write concisely.", aiming to reduce the verbosity of the responses.
This is based on the observation that image clusters are often described by succinct short descriptors, e.g., the datasets in our experiments or ImageNet-based \cite{5206848} clustering datasets. Thus, these prompts align with our knowledge about the clustering tasks.

\textbf{Initial Clustering} First, we perform VQA for each pair of images and prompts, generating responses relevant to the visual content.  Next, we create text representations using both traditional TF-IDF \cite{10.5555/106765.106782} and an advanced sentence embedding model, namely \textit{gte-large} \cite{li2023towards}.  Finally, we apply k-means to these text representations and obtain a clustering for each prompt and each representation.

\subsection{Grouping}

The input to the grouping stage are $n$ pairs of prompt and corresponding clustering $(p_i, \pi_i), i \in [n]$ and the number of ground truth clustering sizes $\{z_1, \dots, z_t\}, z_i \in \mathbb{N}$, e.g. $\{2, 4\}$.
The goal is to obtain groups of clusterings to later find consensus between the individual clusterings explaining the data from a similar perspective. 
This is displayed in step 5 of Figure \ref{fig:main_figure}.
Specifically, we aim to connect semantically similar clusterings and detect potential outlier clusterings, which are caused by prompts leading to unexpected or inconsistent VQA outcomes and are not useful for our final clustering. Find examples of generated text in Table \ref{tab:examples}.

Therefore, we compute the similarity of two clusterings using Adjusted Mutual Information (AMI) \cite{JMLR_v11_vinh10a}. 
We choose AMI as it is a standard clustering metric based on information theory.
Then, we use a spanning-tree-based hierarchical clustering \cite{muellner2011modern} algorithm\footnote{Algorithm is readily available in the Scipy library \cite{2020SciPy-NMeth}: \url{https://docs.scipy.org/doc/scipy/reference/generated/scipy.cluster.hierarchy.fcluster.html}} to systematically group similar prompts, facilitating a structured analysis of clustering behavior (see step 5 of Figure \ref{fig:main_figure}).
The basic idea is that for a threshold $\tau \in (0, 1)$, two clusterings $A, B$ are connected if their distance is less than $\tau$, i.e. $AMI(A,B) < \tau$.
Here, we use two strategies, which we call ``min'' and ``max''. 
For ``min'', we find a minimum threshold such that the resulting number of groups is equal to the number of expected groupings $t$. For ``max'', we find a maximum threshold such that this constraint is fulfilled. We use the trivial solution to iterate over all thresholds in $\tau \in (0, 1)$ in steps of $0.02$ as the runtime is negligible.

\subsection{Aggregation}

In the end, we synthesize each group of clusterings. Given that we aggregate potentially very different clusterings, it is beneficial to use different aggregation schemes.
Therefore, we apply multiple consensus clustering algorithms for each group and choose the instance with the lowest clustering loss. Specifically, we employ MCLA, HBGF, CSPA, and NMF \citep{clusterensembles, nmfconsensus} to aggregate the clusterings within the groups\footnote{We used the library Cluster Ensembles: \url{https://github.com/GGiecold-zz/Cluster_Ensembles}}. 
Thereby, we aim to use consensus clustering to combine the strength of multiple clusterings. In our ablation analysis we also test the simple solution where we concatenate the generated text of the prompts in each clustering group and perform k-means on the concatenated string.

\section{Experiments}

In this section, we introduce the experimental setup. Afterwards, we discuss the main results obtained, highlighting the performance of TGAICC and text-based methods. Additionally, we provide an ablation study, systematically analyzing the impact of various components and prompts on the overall performance. Finally, we perform a simple cluster explainability method to get a textual overview of the data.

\begin{table}[htbp]

\centering
\resizebox{.5\textwidth}{!}{
\begin{tabular}{l|l|l|l}
Dataset & \#samples & \#clusters & Size \\ \hline
Fruits-360 & 4856 & 4; 4 & 100x100 \\
Cards & 8029 & 3; 4 & 224x224 \\
GTSRB & 6720 & 4; 2 & 15x15 to 250x250 \\
NR-Objects & 10000 & 6; 2; 3 & 100x100 \\
\end{tabular}
}

\caption{Overview of statistics of the dataset. The third column contains the number of clusters in the ground truth clusterings.}
\label{tab:dataset_stats}

\end{table}

\subsection{Experimental Setup}
\label{subsec:exp_setup}

In this section, we outline the key components of our experimental setup, including the evaluation metrics, data representations, and models used.  All experiments were run on a single A100 GPU. VQA took approximately 24 hours and TGAICC experiments took about the same time. Embedding text and running consensus clustering are the most time consuming elements.
Each algorithm is executed 10 times with different random states, and we report the average performance across these runs.

\begin{table*}
\centering
\resizebox{\textwidth}{!}{

\begin{tabular}{lll|ccccc|cc|cc|c}
\toprule
 &  &  & \multicolumn{5}{c|}{Image} & \multicolumn{2}{c|}{TF-IDF} & \multicolumn{2}{c|}{SBERT} & TGAICC \\
Dataset & Type &  & k-means & orth-1 & orth-2 & Nr-Kmeans & ENRC & Avg. Prompt & Concatenate & Avg. Prompt & Concatenate &  \\
\midrule
\multirow[c]{4}{*}{Fruits-360} & \multirow[c]{2}{*}{fruit} & ARI & 27.40 & \uline{31.50} & 30.80 & \bfseries 35.40 & 26.00 & 14.80 & 20.10 & 15.10 & 17.20 & 18.60 \\
 &  & AMI & 41.30 & 42.10 & \uline{42.90}& \bfseries 50.60 & 36.70 & 24.80 & 32.20 & 25.00 & 28.60 & 26.90 \\
 & \multirow[c]{2}{*}{colour} & ARI & 33.20 & 35.40 & 33.50 & 40.90 & 39.70 & 40.00 & \uline{54.60}& 47.40 & 51.60 & \bfseries 54.70 \\
 &  & AMI & 47.30 & 53.30 & 51.70 & 55.50 & 54.90 & 50.70 & \bfseries 65.50 & 56.90 & 60.80 & \uline{64.80}\\
 \midrule
\multirow[c]{4}{*}{GTSRB} & \multirow[c]{2}{*}{type} & ARI & 41.70 & 46.80 & 46.80 & 22.70 & 38.20 & 45.10 & \bfseries 61.00 & 49.70 & 57.50 & \uline{58.00}\\
 &  & AMI & 51.50 & 55.50 & 55.50 & 38.60 & \bfseries 72.50 & 52.40 & \uline{67.90}& 55.50 & 63.20 & 64.60 \\
 & \multirow[c]{2}{*}{colour} & ARI & 23.00 & 0.10 & 0.10 & 49.00 & 55.90 & 79.20 & 87.40 & \uline{88.50}& \bfseries 90.00 & 88.00 \\
 &  & AMI & 33.40 & 0.10 & 0.10 & 43.30 & 28.30 & 73.70 & 82.30 & 82.70 & \bfseries 84.50 & \uline{83.00}\\
 \midrule
\multirow[c]{4}{*}{Cards} & \multirow[c]{2}{*}{rank} & ARI & 30.10 & 29.70 & 26.70 & \uline{35.70}& 33.10 & 24.30 & 24.70 & 33.00 & \bfseries 50.70 & 34.70 \\
 &  & AMI & 47.80 & 47.60 & 41.70 & \uline{55.20}& 52.40 & 41.30 & 41.60 & 50.00 & \bfseries 68.40 & 50.20 \\
 & \multirow[c]{2}{*}{suit} & ARI & 25.90 & 1.10 & 3.80 & 10.60 & 14.30 & 19.70 & \uline{29.60}& 25.90 & 28.30 & \bfseries 29.70 \\
 &  & AMI & 34.40 & 1.20 & 8.20 & 16.60 & 19.80 & 27.40 & \uline{37.10} & 33.60 & 35.40 & \bfseries 38.30 \\
 \midrule
\multirow[c]{6}{*}{NR-Objects} & \multirow[c]{2}{*}{shape} & ARI & \uline{95.30}& 94.40 & 94.40 & 76.00 & 72.70 & 65.70 & 94.50 & 75.90 & 95.00 & \bfseries 100.00 \\
 &  & AMI & 96.20 & 95.10 & 95.10 & 82.20 & 82.70 & 71.30 & \uline{96.50}& 79.40 & 95.80 & \bfseries 100.00 \\
 & \multirow[c]{2}{*}{material} & ARI & 0.00 & 26.70 & 30.60 & \uline{30.70}& \bfseries 31.60 & 9.20 & 1.60 & 14.80 & 0.00 & 9.00 \\
 &  & AMI & 0.00 & 25.90 & \uline{38.80}& 32.70 & \bfseries 39.40 & 10.10 & 1.80 & 15.00 & 0.00 & 17.10 \\
 & \multirow[c]{2}{*}{colour} & ARI & 9.70 & 87.00 & 75.10 & 50.40 & 45.70 & 66.80 & \uline{91.10}& 81.20 & 83.70 & \bfseries 97.80 \\
 &  & AMI & 21.70 & 93.00 & 79.00 & 65.70 & 66.00 & 81.20 & \uline{95.30} & 88.30 & 91.40 & \bfseries 97.90 \\
 \midrule
\multirow[c]{2}{*}{Avg.} & \multirow[c]{2}{*}{} & ARI & 31.81 & 39.19 & 37.98 & 39.04 & 39.69 & 40.53 & 51.62 & 47.94 & \uline{52.67} & \bfseries 54.50 \\
 &  & AMI & 41.51 & 45.98 & 45.89 & 48.93 & 50.30 & 48.10 & 57.80 & 54.04 & \uline{58.68}& \bfseries 60.31 \\
\bottomrule
\end{tabular}
}

\caption{Main results table. Best results are in bold, second best results are underlinded.}
\label{tab:main_table}
\end{table*} 

\subsubsection{Metrics} 
We employ two widely used metrics to assess the performance of our clustering models. The Adjusted Rand Index (ARI) \cite{rand1971objective} measures the similarity between the predicted cluster assignments and the ground truth labels, adjusting for chance agreement. The Adjusted Mutual Information (AMI) \cite{JMLR_v11_vinh10a} quantifies the shared information between the predicted clusters and the true labels. We multiply by 100 to increase readability.

\subsubsection{Representations}

We utilize image- and text-based representations to capture different aspects of the data.

\textbf{Image Embeddings:} We utilize the LLaVA-NeXT model \cite{liu2023improvedllava}, which incorporates the image encoder of a frozen CLIP model. This allows us to directly use the image embeddings learned during the contrastive pre-training of CLIP \cite{pmlr-v139-radford21a} for our clustering tasks.

\textbf{Statistical text embeddings:} We employ Term Frequency-Inverse Document Frequency (TF-IDF) embeddings, a standard word frequency-based technique for representing documents.

\textbf{Neural text embeddings:} To better capture semantic relationships, we employ the ``gte-large''\footnote{Model is available on Hugging Face (\url{https://huggingface.co/thenlper/gte-large}, and is used via the Sentence-BERT (SBERT) library \cite{reimers-2019-sentence-bert}} model \cite{li2023towards}, a state-of-the-art sentence encoder.

\subsubsection{Datasets}

In the following, we briefly describe the used datasets. Table \ref{tab:dataset_stats} summarizes the relevant statistics for all datasets. More details about datasets and corresponding prompts are in Appendix \ref{sec:appdx_datasets}.

\noindent
\textbf{Cards} \cite{yao2023augdmc} This dataset is primarily used for classification tasks but contains attributes suitable for clustering based on card suit and rank. 

\noindent
\textbf{Fruits-360} \cite{fruit360} The dataset is composed of images that can be clustered by fruit type (citrus, berries, etc.) and color.

\noindent
\textbf{NR-Objects} \cite{Miklautz_Mautz_Altinigneli_Boehm_Plant_2020} The dataset contains images of objects (e.g., cubes), which can be clustered by shape, material, or color.

\noindent
\textbf{German Traffic Sign Recognition (GTSRB)} \cite{gtsrb}  This dataset contains traffic signs and can be clustered by color and traffic sign type.

\subsubsection{Baselines}

We use multiple image-based alternative clustering baselines and baselines using the generated text. The code is implemented using the ClustPy\footnote{\url{https://github.com/collinleiber/ClustPy}} library \cite{leiber2023benchmarking}. Additional details are given in Appendix \ref{sec:appdx_baselines}.

\textbf{Orth} \cite{orth} iteratively identifies several clusterings by first clustering using PCA (keeping $90\%$ of the variance) in combination with k-means and then creating a new orthogonal feature space. There are two strategies for orthogonalization: \textit{orthogonal clustering} (orth-1) and \textit{clustering in orthogonal subspaces} (orth-2).

\textbf{Nr-Kmeans} \cite{nrkmeans} simultaneously optimizes several clusterings by assigning each clustering result a separate subspace in which k-means is executed.

\textbf{ENRC} \cite{Miklautz_Mautz_Altinigneli_Boehm_Plant_2020} is a deep clustering method that assigns multiple clusterings to a dataset by (soft-)assigning each dimension of the embeddings space to a clustering.

\textbf{Avg. Prompt.} We measure the performance of clustering each text generated per prompt and subsequentially report the average performance.

\textbf{Concat. by Category.} We manually group all prompts together that belong to the same clustering type (e.g., ``rank'' or ``suit''), concatenate all generated text, and cluster it using k-means.

\subsection{Main Experiments}


The results of our main experiments are shown in Table \ref{tab:main_table}. They reveal that, on average, text-based methods, including TGAICC, outperform image-based methods.
Further, we observe that TGAICC, on average, demonstrates superiority over avg. prompting and concatenation baselines.
In addition, we can see that clustering by material in the NR-Objects dataset does not work in the text domain. 
\begin{table}
\resizebox{0.48\textwidth}{!}{
\begin{tabular}{lll|cccc|cccc}
\toprule
 &  &  & \multicolumn{4}{c|}{TF-IDF} & \multicolumn{4}{|c}{SBERT} \\
 &  &  & \multicolumn{2}{c}{selection} & \multicolumn{2}{c|}{consensus} & \multicolumn{2}{|c}{selection} & \multicolumn{2}{c}{consensus} \\
 &  &  & min & max & min & max & min & max & min & max \\
\midrule
\multirow[c]{4}{*}{Fruits-360} & \multirow[c]{2}{*}{fruit} & ARI & \uline{20.80} & \bfseries 24.60 & 17.40 & 19.50 & 19.60 & 17.20 & 15.80 & 18.60 \\
 &  & AMI & \uline{36.60}& \bfseries 39.10 & 29.00 & 32.90 & 33.00 & 30.90 & 23.20 & 26.90 \\
 & \multirow[c]{2}{*}{colour} & ARI & 51.80 & 52.20 & 51.10 & 51.90 & \bfseries 58.60 & \uline{57.20}& 54.30 & 54.70 \\
 &  & AMI & 61.70 & 62.20 & 60.70 & 61.40 & \bfseries 71.50 & \uline{66.40} & 64.90 & 64.80 \\
 \midrule
\multirow[c]{4}{*}{GTSRB} & \multirow[c]{2}{*}{type} & ARI & 52.70 & \bfseries 73.20 & 49.70 & 54.10 & 50.80 & \uline{58.00}& 51.60 & \uline{58.00}\\
 &  & AMI & 60.80 & \bfseries 75.40 & 56.60 & 60.40 & 57.80 & 64.10 & 60.00 & \uline{64.60} \\
 & \multirow[c]{2}{*}{colour} & ARI & 74.70 & 74.00 & 87.10 & 87.30 & 73.10 & 70.20 & \bfseries 88.90 & \uline{88.00}\\
 &  & AMI & 70.70 & 70.10 & 81.60 & 81.80 & 70.20 & 68.60 & \bfseries 83.20 & \uline{83.00}\\
\midrule
\multirow[c]{4}{*}{Cards} & \multirow[c]{2}{*}{rank} & ARI & 28.60 & 28.00 & 27.00 & 26.90 & \uline{40.30}& \bfseries 56.00 & 36.30 & 34.70 \\
 &  & AMI & 48.30 & 47.00 & 47.60 & 46.20 & \uline{58.50}& \bfseries 72.20 & 51.30 & 50.20 \\
 & \multirow[c]{2}{*}{suit} & ARI & 19.40 & 20.10 & 21.60 & 21.10 & 19.60 & 19.90 & \uline{22.40}& \bfseries 29.70 \\
 &  & AMI & 29.30 & 28.90 & 23.60 & 23.70 & \uline{30.50}& 27.30 & 27.40 & \bfseries 38.30 \\
 \midrule
\multirow[c]{6}{*}{NR-Objects} & \multirow[c]{2}{*}{shape} & ARI & 98.70 & 98.90 & \uline{99.30}& \uline{99.30}& \bfseries 100.00 & \bfseries 100.00 & \bfseries 100.00 & \bfseries 100.00 \\
 &  & AMI & 97.60 & 97.90 & \uline{98.90}& 98.70 & \bfseries 100.00 & \bfseries 100.00 & \bfseries 100.00 & \bfseries 100.00 \\
 & \multirow[c]{2}{*}{material} & ARI & \bfseries 23.10 & \uline{22.40}& 0.10 & 1.00 & 0.00 & 0.00 & 9.00 & 9.00 \\
 &  & AMI & \bfseries 22.70 & \uline{22.20}& 0.10 & 1.80 & 0.00 & 0.00 & 17.10 & 17.10 \\
 & \multirow[c]{2}{*}{colour} & ARI & 33.30 & 33.30 & 80.00 & \uline{84.10}& 43.60 & 43.60 & \bfseries 97.80 & \bfseries 97.80 \\
 &  & AMI & 65.20 & 65.20 & 87.50 & \uline{90.10} & 66.60 & 66.60 & \bfseries 97.90 & \bfseries 97.90 \\
 \midrule
\multirow[c]{2}{*}{Avg.} & \multirow[c]{2}{*}{} & ARI & 44.79 & 47.41 & 48.14 & 49.47 & 45.07 & 46.90 & \uline{52.90}& \bfseries 54.50 \\
 &  & AMI & 54.77 & 56.44 & 53.96 & 55.22 & 54.23 & 55.12 & \uline{58.33} & \bfseries 60.31 \\
\bottomrule
\end{tabular}
}
\caption{An ablation analysis of TGAICC, where ``min'' and ``max'' refer to the thresholding strategy, and concatenation and consensus to the aggregation scheme. Consensus-max resembles TGAICC. The best results are in bold, and the second best results are underlined.}
\label{tab:aggregation_ablation}
\end{table}

See Table \ref{tab:examples} for VQA examples. 
The main takeaway is that, in many cases, the VQA model provides too much information, even information that should be used for a different clustering, e.g., color or shape. This highlights a core limitation of our methodology. If the text generation does not work sufficiently well, the subsequent clustering can not work. 
\begin{table*}
    \centering
\resizebox{\textwidth}{!}{
\begin{tabular}{ll|rr|rr}
\toprule
 &  & \multicolumn{2}{|c|}{TF-IDF}& \multicolumn{2}{|c}{SBERT}\\
 & Prompt & ARI & AMI & ARI & AMI  \\
\midrule
\multirow[c]{6}{*}{suit} & Can you tell me the suit of the playing card shown in the picture? & 25.42 & 31.25 & 25.42 & 31.25 \\
 & What suit does the playing card in the image belong to? & 25.59 & 33.53 & 25.59 & 33.53 \\
 & Could you identify the suit of the playing card depicted in the photo? & \bfseries 29.29 & 33.64 & \bfseries 29.29 & 33.64 \\
 & Can you tell me the suit of the playing card shown in the picture? Answer concisely. & 24.25 & \bfseries 37.32 & 24.25 & \bfseries 37.32 \\
 & What suit does the playing card in the image belong to? Answer concisely. & \uline{28.97} & \uline{36.35} & \uline{28.97} & \uline{36.35} \\
 & Could you identify the suit of the playing card depicted in the photo? Answer concisely. & 21.85 & 29.71 & 21.85 & 29.71 \\
 \midrule
\multirow[c]{6}{*}{rank} & Can you tell me the rank of the card shown in the picture? & 26.76 & 43.33 & 26.76 & 43.33 \\
& What is the numerical or face value of the card displayed in the image? & 32.06 & 47.36 & 32.06 & 47.36 \\
 & What level or position does the card in the photo hold? & 31.52 & 47.28 & 31.52 & 47.28 \\
 & Can you tell me the rank of the card shown in the picture? Answer concisely. & \uline{37.48} & \uline{55.42} & \uline{37.48} & \uline{55.42} \\
 & What is the numerical or face value of the card displayed in the image? Answer concisely. & \bfseries 38.79 & \bfseries 56.09 & \bfseries 38.79 & \bfseries 56.09 \\
 & What level or position does the card in the photo hold? Answer concisely. & 31.15 & 50.44 & 31.15 & 50.44 \\
\bottomrule
\end{tabular}
}
\caption{Ablation study comparing the clustering performance of individual prompts. Here we show a case study based on the Cards dataset. The best results are in bold, and the second-best results are underlined.}
    \label{tab:per_prompt}
\end{table*}
Nevertheless, TGAICC is model-agnostic and can be used with any VQA image-to-text system.
In this way, it can make use of future advancements in VQA models.

\subsection{Aggregation ablation}

In this ablation study, we investigate the aggregation components of TGAICC. More specifically, we investigate the impact of the thresholding and aggregation strategies on clustering performance.

\textbf{Setup. } We ablate the ``min'' and ``max'' thresholding strategies, which find the minimum and maximum threshold such that the number of clustering groups corresponds to the expected number of alternative clusterings.
We experiment with the consensus-clustering-based aggregation scheme used in TGAICC and compare it to the simple ``concatenation'' baseline, which concatenates the text of the corresponding clustering groups. Results are shown in Table \ref{tab:aggregation_ablation}. Note that TGAICC is consensus-max.

\textbf{Results.} Our analysis reveals that consensus clustering outperforms concatenation-based selection. Furthermore, SBERT-based clustering outperforms TF-IDF-based clustering. 
We observe that the performance of the 'min' and the 'max' strategy are very similar, which indicates the stability of the method w.r.t. the thresholding strategy.

\subsection{Individual prompt analysis}

TGAICC is based on the aggregation of multiple clusterings, which in turn are based on generated texts using different VQA prompts. 
As known from other tasks \cite{sclar2024quantifying}, different prompts potentially result in high-performance variance.

\textbf{Setup.} In Table \ref{tab:per_prompt} we analyze the clustering performance per prompt on the case study of the Cards dataset. Note that again we execute k-means 10 times and present the averaged results. 

\textbf{Results.} In the case study, the addition of the prompt ``Answer concisely'' mostly yields similar clustering results, with a slight performance advantage when the ``Answer concisely'' prompt is included. 
Furthermore, we note a significant variation in clustering performance across different prompts. 

\subsection{Explainability}

Our cluster explainability aims to provide insights into the different clustering possibilities. This understanding is essential for interpreting and validating clustering outcomes. We use a simple word frequency-based algorithm to explain the data.

\begin{table}
    \centering
    
\resizebox{.45\textwidth}{!}{
\begin{tabular}{c|c|c|c}
\toprule
\multicolumn{2}{c|}{Suit} & \multicolumn{2}{c}{Rank} \\ 
Truth &Top Words &  Truth &Top Words \\
\midrule
heart  & heart &ace &ace \\
diamond & diamond &king &king \\
club & club & queen &queen \\
spade & spade & jack &jack \\
& &5 &heart \\
& &9 &spade \\
& &3 &rank \\
& &4 &club \\
& &6 &diamond \\
& &10 &10 \\
& &2 &twos \\
& &8 &8 \\
& &7 &7 \\
\bottomrule
\end{tabular}
}
\caption{This table shows how we are able to explain the datasets by listing the top most used words of the two final clusterings. For each top word, we show the ideal ground truth cluster name assignment.}
\label{tab:general_explainability}
\end{table}

\begin{table*}
\centering
\resizebox{\textwidth}{!}{
\begin{tabular}{l|c|p{0.43\textwidth}|p{0.52\textwidth}}
\toprule
  \textbf{Material} & \textbf{Image} & \textbf{Question} & \textbf{Generated Text} \\
\midrule
  Rubber & \multirow{2}{*}{\adjustbox{valign=t}{\includegraphics[width=2.5cm]{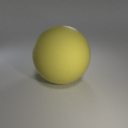}}} & What substance is the item in the picture made of? Answer concisely. & Plastic \\
   \cline{3-4} 
   & & What substance is the item in the picture made of? & The item in the picture appears to be a sphere made of a smooth, shiny material that could be plastic, metal, or a similar synthetic material. It's difficult to determine the exact material without more information or a closer inspection. \\ \hline
  Metal & \multirow{3}{*}{\adjustbox{valign=t}{\includegraphics[width=2.5cm]{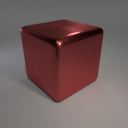}}}&  What substance is the item in the picture made of? Answer concisely. & The item in the picture is made of metal. \\
   \cline{3-4}  
   & & Can you identify the material used in the object shown in the image? Answer concisely. & The object in the image appears to be made of a shiny, reflective material, possibly metal or a metal-like material. \\
    \cline{3-4} 
   & & What is the composition of the object depicted in the photo? Answer concisely. & The object in the photo is a 3D rendering of a red cube. \\
\bottomrule
\end{tabular}
} 
\caption{Some VQA examples on the NR-Objects dataset. While all answers are reasonable, they often provide too much information, such as information about the shape, or make statements about the ambiguity of the underlying material. } 
\label{tab:examples}
\end{table*}

\textbf{Setup.} For a specific final clustering of TGAICC, we concatenate all generated texts belonging to the prompts used to obtain this clustering. The resulting text is changed to lowercase and made singular.
Finally, to explain a final clusterings, we determine the $z$ most frequently occurring words, where $z$ is the number of clusters of the respective clustering. For instance for the suit clustering $z=4$. Table \ref{tab:general_explainability} shows the resulting words for the Cards dataset. We reorder the ground truth cluster names suitably.

\textbf{Analysis.}  Notably, the explainability method effectively identifies the ``suits'' cluster names, providing a comprehensive description of this clustering type, even though clustering performance has an AMI of less than 40\%.
Additionally, the frequency analysis exposed many of the card types in the dataset.
However, suit names are also assigned as cluster names for the expected card ranking clusters (e.g., ``heart'' as the top word of the ``10'' cluster). 
Figure \ref{tab:examples} presents concrete examples demonstrating that Visual Question Answering (VQA) models often provide additional information, such as suit, thereby explaining the inclusion of suits as rank names.

\section{Discussion} 

\subsection{Text-driven data interaction} 

Textual data, as a fundamental form of human communication, offers a natural and intuitive interface for interacting with complex datasets. Our method capitalizes on this inherent connection by utilizing textual prompts for VQA models to guide alternative clusterings. This approach aligns with real-world scenarios where users possess domain knowledge and seek answers to specific questions.  
We envision a future where users can explore datasets from diverse perspectives and test emerging hypotheses interactively, using text. Our research contributes to this scenario.




\subsection{Domain Expertise} 

Our approach incorporates domain expertise, recognizing that users often either have specific questions or some knowledge about their data. This stands in contrast to the traditional clustering setup, which typically operates without user input. By leveraging domain knowledge, our approach aligns with real-world scenarios and allows for more targeted and insightful data exploration.





\section{Conclusion}

In conclusion, this research introduces TGAICC (Text-Guided Alternative Image Consensus Clustering), a novel approach that leverages prompting to inject domain knowledge and human intuition into the clustering process. The experiments on four common alternative image clustering benchmarks demonstrate that TGAICC outperforms competitive image- and text-based baselines.
Furthermore, the inherent explainability of text enables a deeper understanding of the underlying data cluster formations. 

By utilizing textual prompts, we can explicitly guide the clustering process from various angles simultaneously, aligning with human intuition. 
This approach offers a more comprehensive and flexible way to analyze visual data, revealing insights that might be missed by traditional clustering methods.


\bibliography{custom}
\bibliographystyle{acl_natbib}

\appendix

\section{Example Appendix}
\label{sec:appendix}

\section{Datasets}
\label{sec:appdx_datasets}
In addition to the dataset description presented in Section \ref{subsec:exp_setup}, we provide the following supplementary materials to enhance the reader's understanding. 
Table \ref{tab:appdx_cluster_names_prompts} shows examples of images from each dataset.
Furthermore, Table \ref{tab:appdx_cluster_names_prompts} provides all prompts generated by GPT-4, paired with their corresponding ground truth cluster names for each clustering type. 
Together, they give a good insight into the datasets and a textual interaction with them.

\begin{table*}
\centering
\resizebox{\textwidth}{!}{
\begin{tabular}{l|l|l|l}
\toprule
Dataset & Type & Cluster Names & Prompts \\
\midrule
\multirow[t]{12}{*}{Fruits-360} & \multirow[t]{6}{*}{fruit} & \multirow[t]{6}{*}{apple, banana, cherry, grape} & What kind of produce is shown in the picture? \\
 &  &  & Can you identify the type of produce depicted in the image? \\
 &  &  & What category of produce does the image represent? \\
 &  &  & What kind of produce is shown in the picture? Answer concisely. \\
 &  &  & Can you identify the type of produce depicted in the image? Answer concisely. \\
 &  &  & What category of produce does the image represent? Answer concisely. \\
\cline{2-4} \cline{3-4}
 & \multirow[t]{6}{*}{colour} & \multirow[t]{6}{*}{burgundy, green, red, yellow} & Can you tell me the color of the fruits and vegetables shown in the picture? \\
 &  &  & What color is the produce displayed in the photo? \\
 &  &  & What hue are the items in the picture? \\
 &  &  & Can you tell me the color of the fruits and vegetables shown in the picture? Answer concisely. \\
 &  &  & What color is the produce displayed in the photo? Answer concisely. \\
 &  &  & What hue are the items in the picture? Answer concisely. \\
\midrule
\multirow[t]{12}{*}{GTSRB} & \multirow[t]{6}{*}{type} & \multirow[t]{6}{*}{70\_limit, dont\_overtake, go\_right, go\_straight} & What kind of traffic sign is shown in the picture? \\
 &  &  & Can you identify the category of the traffic sign displayed in the image? \\
 &  &  & What class of traffic sign is depicted in the photo? \\
 &  &  & What kind of traffic sign is shown in the picture? Answer concisely. \\
 &  &  & Can you identify the category of the traffic sign displayed in the image? Answer concisely. \\
 &  &  & What class of traffic sign is depicted in the photo? Answer concisely. \\
\cline{2-4} \cline{3-4}
 & \multirow[t]{6}{*}{colour} & \multirow[t]{6}{*}{blue, red} & What color is the traffic sign shown in the picture? \\
 &  &  & Can you tell me the color of the traffic sign depicted in the image? \\
 &  &  & What hue is the traffic sign in the photograph? \\
 &  &  & What color is the traffic sign shown in the picture? Answer concisely. \\
 &  &  & Can you tell me the color of the traffic sign depicted in the image? Answer concisely. \\
 &  &  & What hue is the traffic sign in the photograph? Answer concisely. \\
\midrule
\multirow[t]{18}{*}{NR-Objects} & \multirow[t]{6}{*}{shape} & \multirow[t]{6}{*}{cube, cylinder, sphere} & Can you identify the form of the object shown in the picture? \\
 &  &  & What form does the object in the picture take? \\
 &  &  & Could you tell me the configuration of the object depicted in the image? \\
 &  &  & Can you identify the form of the object shown in the picture? Answer concisely. \\
 &  &  & What form does the object in the picture take? Answer concisely. \\
 &  &  & Could you tell me the configuration of the object depicted in the image? Answer concisely. \\
\cline{2-4} \cline{3-4}
 & \multirow[t]{6}{*}{material} & \multirow[t]{6}{*}{metal, rubber} & What substance is the item in the picture made of? \\
 &  &  & Can you identify the material used in the object shown in the image? \\
 &  &  & What is the composition of the object depicted in the photo? \\
 &  &  & What substance is the item in the picture made of? Answer concisely. \\
 &  &  & Can you identify the material used in the object shown in the image? Answer concisely. \\
 &  &  & What is the composition of the object depicted in the photo? Answer concisely. \\
\cline{2-4} \cline{3-4}
 & \multirow[t]{6}{*}{colour} & \multirow[t]{6}{*}{blue, gray, green, purple, red, yellow} & What color is the item shown in the picture? \\
 &  &  & Can you tell me the color of the object depicted in the image? \\
 &  &  & What hue does the object in the photo have? \\
 &  &  & What color is the item shown in the picture? Answer concisely. \\
 &  &  & Can you tell me the color of the object depicted in the image? Answer concisely. \\
 &  &  & What hue does the object in the photo have? Answer concisely. \\
\cline{1-4} \cline{2-4} \cline{3-4}
\multirow[t]{12}{*}{Cards} & \multirow[t]{6}{*}{rank} & \multirow[t]{6}{*}{ace, eight, five, four, jack, king, nine, queen, seven, six, ten, three, two} & Can you tell me the rank of the card shown in the picture? \\
 &  &  & What is the numerical or face value of the card displayed in the image? \\
 &  &  & What level or position does the card in the photo hold? \\
 &  &  & Can you tell me the rank of the card shown in the picture? Answer concisely. \\
 &  &  & What is the numerical or face value of the card displayed in the image? Answer concisely. \\
 &  &  & What level or position does the card in the photo hold? Answer concisely. \\
\cline{2-4} \cline{3-4}
 & \multirow[t]{6}{*}{suit} & \multirow[t]{6}{*}{clubs, diamonds, hearts, spades} & Can you tell me the suit of the playing card shown in the picture? \\
 &  &  & What suit does the playing card in the image belong to? \\
 &  &  & Could you identify the suit of the playing card depicted in the photo? \\
 &  &  & Can you tell me the suit of the playing card shown in the picture? Answer concisely. \\
 &  &  & What suit does the playing card in the image belong to? Answer concisely. \\
 &  &  & Could you identify the suit of the playing card depicted in the photo? Answer concisely. \\
\bottomrule
\end{tabular}
}
\caption{Overview of the datasets, the names of their ground truth clusterings, and all generated prompts.}
\label{tab:appdx_cluster_names_prompts}
\end{table*}

\begin{table}
    \centering
    \resizebox{0.9\textwidth}{!}{
\begin{tabular}{p{0.2\textwidth} |l|l|l}
\toprule
\textbf{dataset} & \textbf{Image 1} & \textbf{Image 2} & \textbf{Image 3} \\
\midrule
Fruits-360 & \includegraphics{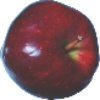} & \includegraphics{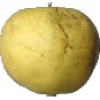} & \includegraphics{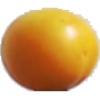} \\
\midrule
GTSRB & 
\includegraphics{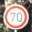} & \includegraphics{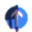} & \includegraphics{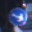} \\
\midrule
NR-Objects & \includegraphics{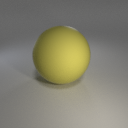} & \includegraphics{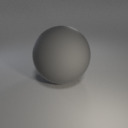} & \includegraphics{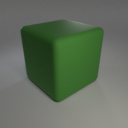} \\
\midrule
Cards & \includegraphics{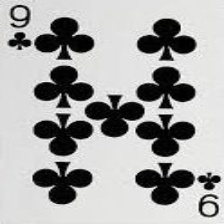} & \includegraphics{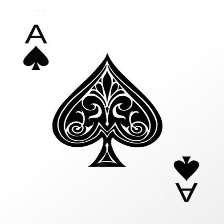} & \includegraphics{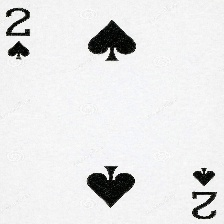} \\
\bottomrule
\end{tabular}
}
\caption{A few example images for each dataset.}
\label{tab:example_images}
\end{table}

\section{Baselines}
\label{sec:appdx_baselines}

In the following, additional details for the baselines are given. We employ all of the following ones on the image embedding of the CLIP encoder of LLaVA-NeXT:

\textbf{K-means. } For all k-means runs, we utilize the k-means++ initialization strategy and set the number of initializations to 1. The code was implemented using scikit-learn \cite{scikit-learn}.

\textbf{Nr-Kmeans. } We set a limit of 300 maximum iterations.

\textbf{Orth 1/2. } We set the explained variance parameter to 90\%.

\textbf{ENRC. } We try the learning rates to $lr=0.001, 0.0001$, use NR-Kmeans as initialization and batch size to $128$, run the autoencoder for $200$ epochs.

\end{document}